\documentclass[11pt,a4paper]{article}
\usepackage[hyperref]{acl2020}
\usepackage{times}
\usepackage{latexsym}

\usepackage{comment}
\usepackage{tikz-dependency}
\usepackage{adjustbox}
\usepackage{amsfonts}
\usepackage{amsmath}
\usepackage{multirow}

\usepackage{url}

\usepackage{url}

\usepackage{microtype}

\aclfinalcopy 

\title{Improving Slot Filling by Utilizing Contextual Information}

\author{
Amir Pouran Ben Veyseh\textsuperscript{*}\thanks{* This work was done when the first author was an intern at Adobe Research.}\textsuperscript{\rm 1}, Franck Dernoncourt\textsuperscript{\rm 2}, \\
{\bf and Thien Huu Nguyen\textsuperscript{\rm 1}} \\
\textsuperscript{\rm 1}Department of Computer and Information Science, \\ University of Oregon, Eugene, Oregon, USA\\
\textsuperscript{\rm 2}Adobe Research, San Jose, CA, USA\\
{\tt \{apouranb, thien\}@cs.uoregon.edu} \\
{\tt franck.dernoncourt@adobe.com}
}

\renewcommand\footnotemark{}

\date{}

\begin{document}
\maketitle
\begin{abstract}
Slot Filling (SF) is one of the sub-tasks of Spoken Language Understanding (SLU) which aims to extract semantic constituents from a given natural language utterance. It is formulated as a sequence labeling task. Recently, it has been shown that contextual information is vital for this task. However, existing models employ contextual information in a restricted manner, e.g., using self-attention.
Such methods fail to distinguish the effects of the context on the word representation and the word label. To address this issue, in this paper, we propose a novel method to incorporate the contextual information in two different levels, i.e., representation level and task-specific (i.e., label) level. Our extensive experiments on three benchmark datasets on SF show the effectiveness of our model leading to new state-of-the-art results on all three benchmark datasets for the task of SF.

\end{abstract}

\section{Introduction}

Slot Filling (SF) is the task of identifying the semantic constituents expressed in natural language utterance. It is one of the sub-tasks of spoken language understanding (SLU) and plays a vital role in personal assistant tools such as Siri, Alexa, and Goolge Assistant. This task is formulated as a sequence labeling problem. For instance, in the given sentence ``\textit{Play Signe Anderson chant music that is newest.}'', the goal is to identify ``\textit{Signe Anderson}'' as ``\textit{artist}'', ``\textit{chant music}'' as ``\textit{music-item}'' and ``\textit{newest}'' as ``\textit{sort}''.

Early work on SF has employed feature engineering methods to train statistical models, e.g., Conditional Random Field \cite{raymond:07}. Later, deep learning emerged as a promising approach for SF \cite{yao:14,Peng:15,liu:16}. The success of deep models could be attributed to pre-trained word embeddings to generalize words and deep learning architectures to compose the word embeddings to induce effective representations. In addition to improving word representation using deep models, \cite{liu:16} showed that incorporating the context of each word into its representation could improve the results. Concretely, the effect of using context in word representation is two-fold: (1) \textbf{Representation Level}: As the meaning of the word is dependent on its context, incorporating the contextual information is vital to represent the true meaning of the word in the sentence (2) \textbf{Task Level}: For SF, the label of the word is related to the other words in the sentence and providing information about the other words, in prediction layer, could improve the performance. Unfortunately, the existing work employs the context in a restricted manner, e.g., via attention mechanism, which obfuscates the model about the two aforementioned effects of the contextual information. 

In order to address the limitations of the prior work to exploit the context for SF, in this paper, we propose a multi-task setting to train the model. More specifically, our model is encouraged to explicitly ensure the two aforementioned effects of the contextual information for the task of SF. In particular, in addition to the main sequence labeling task, we introduce new sub-tasks to ensure each effect. Firstly, in the representation level, we enforce the consistency between the word representations and its context. This enforcement is achieved via increasing the Mutual Information (MI) between these two representations. Secondly, in the task level, we propose two new sub-tasks: (1) To predict the label of the word solely from its context and (2) To predict which labels exist in the given sentence in a multi-label classification setting. By doing so, we encourage the model to encode task-specific features in the context of each word. Our extensive experiments on three benchmark datasets, empirically prove the effectiveness of the proposed model leading to new the state-of-the-art results on all three datasets.

\section{Related Work}

In the literature, Slot Filling (SF), is categorized as one of the sub-tasks of spoken language understanding (SLU). Early work employed feature engineering for statistical models, e.g., Conditional Random Field \cite{raymond:07}. Due to the lack of generalisation ability of feature based models, deep learning based models superseded them \cite{yao:14,Peng:15,kurata:16,hakkani:16}. Also, joint models to simultaneously predict the intent of the utterance and to extract the semantic slots has also gained a lot of attention \cite{guo:14,liu:16,zhang:16,wang:18,goo:18,qin:19,e:19}. In addition to the supervised setting, recently other setting such as progressive learning \cite{shen:19} or zero-shot learning has also been studied \cite{shah:19}. To the best of our knowledge, none of the existing work introduces a multi-task learning solely for the SF to incorporate the contextual information in both representation and task levels.

\section{Model}

Our model is trained in a multi-task setting in which the main task is slot filling to identify the best possible sequence of labels for the given sentence. In the first auxiliary task we aim to increase consistency between the word representation and its context. The second auxiliary task is to enhance task specific information in contextual information. In this section, we explain each of these tasks in more details.

\subsection{Slot Filling}
Formally, the input to SF model is a sequence of words $X=[x_1,x_2,\ldots,x_n]$ and our goal is to predict the sequence of labels $Y=[y_1,y_2,\ldots,y_n]$. In our model, the word $x_i$ is represented by vector $e_i$ which is the concatenation of the pre-trained word embedding and POS tag embedding of the word $x_i$. In order to obtain a more abstract representation of the words, we employ a Bi-directional Long Short-Term Memory (BiLSTM) over the word representations $E=[e_1,e_2,\ldots,e_n]$ to generate the abstract vectors $H=[h_1,h_2,\ldots,h_n]$. The vector $h_i$ is the final representation of the word $x_i$ and is fed into a two-layer feed forward neural net to compute the label scores $s_i$ for the given word, $s_i = FF(h_i)$. As the task of SF is formulated as a sequence labeling task, we exploit a conditional random field (CRF) layer as the final layer of SF prediction. More specifically, the predicted label scores $S=[s_1,s_2,\ldots,s_n]$ are provided as emission score to the CRF layer to predict the label sequence $\hat{Y}=[\hat{y}_1,\hat{y}_2,\ldots,\hat{y}_n]$. 
To train the model, the negative log-likelihood is used as the loss function for SF prediction, i.e., $\mathcal{L}_{pred}$.

\subsection{Consistency between Word and Context}
In this sub-task we aim to increase the consistency between the word representation and its context. To obtain the context of each word, we use max pooling over the outputs of the BiLSTM for all words of the sentence excluding the word itself, $ h^c_i = Max\-Pooling({h_1,h_2,...,h_n}/h_i)$. We aim to increase the consistency between vectors $h_i$ and $h^c_i$. To this end, we propose to maximize the Mutual Information (MI) between the word representation and its context. In information theory, MI evaluates how much information we know about one random variable if the value of another variable is revealed. Formally, the mutual information between two random variable $X_1$ and $X_2$ is obtained by:

\begin{equation}
\begin{split}
    MI(X_1,X_2) = \int_{X_1} \int_{X_2} P(X_1,X_2) * \\ log \frac{P(X_1,X_2)}{P(X_1)P(X_2)} dX_1dX_2
\end{split}
\end{equation}

Using this definition of MI, we can reformulate the MI equation as KL-Divergence between the joint distribution $P_{X_1X_2}=P(X_1,X_2)$ and the product of marginal distributions $P_{X_1\bigotimes X_2}=P(X_1)P(X_2)$:

\begin{equation}
    \label{eq:mi-kl}
    MI(X_1,X_2) = D_{KL}(P_{X_1X_2}||P_{X_1\bigotimes X_2})
\end{equation}

 Based on this understanding of MI, if the two random variables are dependent then the mutual information between them (i.e. the KL-Divergence in equation \ref{eq:mi-kl}) would be the highest. Consequently, if the representations $h_i$ and $h^c_i$ are encouraged to have large mutual information, we expect them to share more information.
 
 Computing the KL-Divergence in equation \ref{eq:mi-kl} could be prohibitively expensive \cite{belghazi:18}, so we need to estimate it. To this end, we exploit the adversarial method introduced in \cite{hjelm:19}. In this method, a discriminator is employed to distinguish between samples from the joint distribution and the product of the marginal distributions to estimate the KL-Divergence in equation \ref{eq:mi-kl}. In our case, the sample from joint distribution is the concatenation $[h_i:h_i^c]$ and the sample from the product of the marginal distribution is the concatenation $[h_i:h_j^c]$ where $h_j^c$ is a context vector randomly chosen from the words in the mini-batch. Formally:
 
\begin{equation}
\label{eq:disc}
\begin{split}
    \mathcal{L}_{disc} = \frac{1}{n}\Sigma_{i=1}^n  -(log(D[h,h^c_i])+ \\ log(1-D([h_i,h^c_j])))
\end{split}
\end{equation} 
Where $D$ is the discriminator. This loss is added to the final loss function of the model.

\subsection{Prediction by Contextual Information}
In addition to increasing consistency between the word representation and its context representation, we aim to increase the task-specific information in contextual representations. To this end, we train the model on two auxiliary tasks. The first one aims to use the context of each word to predict the label of that word. The goal of the second auxiliary task is to use the global context information to predict sentence level labels. We describe each of these tasks in more details in the following subsections.



\subsubsection*{Predicting Word Label}
In this sub-task, we use the context representations of each word to predict its label. It will increase the information encoded in the context of the word about the label of the word. We use the same context vector $h^c_i$ for the $i$-th word as described in the previous section. This vector is fed into a two-layer feed forward neural network with a softmax layer at the end to output the probabilities for each class, $P_i(.|\{x_1,x_2,...,x_n\}/x_i) = softmax(FF(h^c_i))$. Finally, we use the following negative log-likelihood as the loss function to be optimized during training:

\begin{equation}
    \mathcal{L}_{wp} = \frac{1}{n}\Sigma_{i=1}^{n} -log( P_i(y_i|\{x_1,x_2,...,x_n\}/x_i))
\end{equation}



\subsubsection*{Predicting Sentence Labels}
The word label prediction enforces the context of each word to contain information about its label but it lacks a global view about the entire sentence. In order to increase the global information about the sentence in the representation of the words, we aim to predict the labels existing in a sentence from the representations of its words. More specifically, we introduce a new sub-task to predict which labels exit in the given sentence. We formulate this task as a multi-label classification problem. Formally, for each sentence, we predict the binary vector $Y^s=[y^s_1,y^s_2,...,y^s_{|L|}]$ where $L$ is the set of all possible word labels. In the vector $Y^s$, $y^s_i$ is 1 if the sentence $X$ contains $i$-th label from the label set $L$ otherwise it is 0.

To predict vector $Y^s$, we first compute the representation of the sentence. This representation is obtained by max pooling over the outputs of the BiLSTM, $H=Max\-Pooling(h_1,h_2,...,h_n)$. Afterwards, the vector $H$ is fed into a two-layer feed forward neural net with a sigmoid activation function at the end to compute the probability distribution of $Y^s$(i.e., $P_k(.|x_1,x_2,...,x_n) = \sigma_k(FF(H))$ for $k$-th label in $L$). Note that since this task is a multi-label classification, the number of neurons at the final layer is equal to $|L|$. We optimize the following binary cross-entropy loss:

\begin{equation}
\begin{split}
    \mathcal{L}_{sp} = \frac{1}{|L|} \Sigma_{k=1}^{|L|} -(y^s_k * log(P_k(y^s_k|x_1,x_2,...,x_n)) + \\ (1-y^s_k)*log(1-P_k(y^s_k|x_1,x_2,...,x_n)))
\end{split}
\end{equation}
Finally, to train the entire model we optimize the following combined loss:

\begin{equation}
    \mathcal{L} = \mathcal{L}_{pred}+\alpha \mathcal{L}_{discr} + \beta \mathcal{L}_{wp} + \gamma \mathcal{L}_{sp}
\end{equation}
where $\alpha$, $\beta$ and $\gamma$ are the trade-off parameters to be tuned based on the development set performance.

\section{Experiments}

\subsection{Dataset and Parameters}
We evaluate our model on three SF datasets. Namely, we employ ATIS \cite{hemphill:90}, SNIPS \cite{coucke:18} and EditMe \cite{manuvinakurike2018edit}. ATIS and SNIPS are two widely adopted SF dataset and EditMe is a SF dataset for editing images with four slot labels (i.e., \textit{Action}, \textit{Object}, \textit{Attribute} and \textit{Value}). The statistics of the datasets are presented in the Appendix \ref{app:dataset}. Based on the experiments on EditMe development set, the following parameters are selected: GloVe embedding with 300 dimensions to initialize word embedding
; 200 dimensions for the all hidden layers in LSTM and feed forward neural net; 0.1 for trade-off parameters $\alpha$, $\beta$ and $\gamma$; 
and Adam optimizer with learning rate 0.001. Following previous work, we use F1-score to evaluate the model.



\subsection{Baselines}
We compare our model with other deep learning based models for SF. Namely, we compare the proposed model with Joint Seq \cite{hakkani:16}, Attention-Based \cite{liu:16}, Sloted-Gated \cite{goo:18}, SF-ID \cite{e:19}, CAPSULE-NLU \cite{zhang:19}, and SPTID \cite{qin:19}. Note that we compare our model with the single-task version of these baselines. We also compare our model with other sequence labeling models which are not specifically proposed for SF. Namely, we compare the model with CVT\cite{clark:18} and GCDT\cite{Liu:19}. CVT aims to improve input representation using improving partial views and GCDT exploits contextual information to enhance word representations via concatenation of context and word representation. 

\subsection{Results}
Table \ref{tab:results} reports the performance of the model and baselines. The proposed model outperforms all baselines in all datasets. Among all baselines, GCDT achieves best results on two out of three datasets. This superiority shows the importance of explicitly incorporating the contextual information into word representation for SF. However, the proposed model improve the performance substantially on all datasets by explicitly encouraging the consistency between word and its context in representation level and task-specific (i.e., label) level. Also, Table \ref{tab:results} shows that EditMe dataset is more challenging than the other datasets, despite fewer slot types it has. This difficulty could be addressed by the limited number of training examples and more diversity in sentence structures in this dataset.

\begin{table}
  \centering
  \resizebox{.45\textwidth}{!}{
\begin{tabular}{l|c|c|c}
    \textbf{Model} & \textbf{SNIPS} & \textbf{ATIS} & \textbf{EditMe} \\ \hline
    Joint Seq\shortcite{hakkani:16} & 87.3 & 94.3 & -  \\
    Attention-Based\shortcite{liu:16} & 87.8 & 94.2 & - \\
    Sloted-Gated\shortcite{goo:18} & 89.2 & 95.4 & 84.9 \\
    SF-ID\shortcite{e:19} & 90.9 & 95.5 & 85.2 \\
    CAPSULE-NLU\shortcite{zhang:19} & 91.8 & 95.2 & 84.6\\
    SPTID\shortcite{qin:19} & 90.8 & 95.1 & 85.3 \\ \hline
    CVT\shortcite{clark:18} & 91.4 & 94.8 & 85.4 \\
    GCDT\shortcite{Liu:19} & 92.0 & 95.1 & 85.6 \\ \hline
    \textbf{Ours} & \textbf{93.6} & \textbf{95.8} & \textbf{87.2}
\end{tabular}
}
    \caption{Performance of the model and baselines on the Test sets.}
    \label{tab:results}
\end{table}


\subsection{Ablation Study}
Our model consists of three major components: (1) \textbf{MI}: Increasing mutual information between word and its context representations (2) \textbf{WP}: Predicting the label of the word using its context to increase word level task-specific information in the word context (3) \textbf{SP}: Predicting which labels exist in the given sentence in a multi-label classification to increase sentence level task-specific information in word representations. In order to analyze the contribution of each of these components, we also evaluate the model performance when we remove one of the components and retrain the model. The results are reported in Table \ref{tab:ablation}. This Table shows that all components are required for the model to have its best performance. Among all components, the word level prediction using the contextual information has the major contribution to the model performance. This fact shows that contextual information trained to be informative about the final task is necessary to obtain the representations which could boost the performance.


\begin{table}
  \centering
  \resizebox{.39\textwidth}{!}{
\begin{tabular}{l|c|c|c}
    \textbf{Model} & \textbf{SNIPS} & \textbf{ATIS} & \textbf{EditMe} \\ \hline
    \textbf{Full} & \textbf{93.6} & \textbf{95.8} & \textbf{87.2}  \\
    Full - MI & 92.9 & 95.3 & 84.2 \\
    Full - WP & 91.7 & 94.9 & 83.2 \\
    Full - SP & 92.5 & 95.2 & 84.1 \\
\end{tabular}
}
    \caption{Test F1-score for the ablated models}
    \label{tab:ablation}
\end{table}

\section{Conclusion}

In this work, we introduced a new deep model for the task of Slot Filling (SF). In a multi-task setting, our model increases the mutual information between the word representation and its context, improves label information in the context and predicts which concepts are expressed in the given sentence. Our experiments on three benchmark datasets show the effectiveness of our model by achieving the state-of-the-art results on all datasets for the SF task.

\bibliography{acl2020}
\bibliographystyle{acl_natbib}

\clearpage

\appendix

\section{Dataset Statistics}
\label{app:dataset}
In our experiments, we employ three benchmark datasets, ATIS, SNIPS and EditMe. Table \ref{tab:statistics} presents the statistics of these three datasets. Moreover, in order to provide more insight into the EditMe dataset, we report the labels statistics of this dataset in Table \ref{tab:label-statistics}.

\begin{table}[h]
  \centering
\begin{tabular}{c|c|c|c}
    \textbf{Dataset} & \textbf{Train} & \textbf{Dev} & \textbf{Test}  \\
    \hline
    SNIPS & 13,084 & 700 & 700  \\
    ATIS & 4,478 & 500 & 893 \\
    EditMe & 1,737 & 497 & 559 \\
\end{tabular}
    \caption{Total number of examples in test/dev/train splits of the datasets}
    \label{tab:statistics}
\end{table}

\begin{table}[h]
  \centering
\begin{tabular}{c|c|c|c}
    \textbf{Label} & \textbf{Train} & \textbf{Dev} & \textbf{Test}  \\
    \hline
    Action & 1,562 & 448 & 479  \\
    Object & 4,676 & 1,447 & 1,501 \\
    Attribute & 1,437 & 403 & 462 \\
    Value & 507 & 207 & 155 \\
\end{tabular}
    \caption{Label Statistics of EditMe dataset}
    \label{tab:label-statistics}
\end{table}

\end{document}